%% file: iclr2026_main.tex
\documentclass{article} 
\usepackage{iclr2026_conference,times}

\newif\ifhtmlout \htmloutfalse
\ifdefined\pdfsavepos\else\htmlouttrue\fi
\makeatletter\@ifundefined{iclrfinalcopy}{\htmlouttrue}{}\makeatother
\IfFileExists{LaTeXML.sty}{\htmlouttrue}{}

\usepackage{hyperref}
\usepackage{url}

\usepackage[]{mdframed}
\usepackage[utf8]{inputenc}
\usepackage{amsthm}
\usepackage{dblfloatfix}
\input{math_commands.tex}

\usepackage{listings}
\usepackage{hyperref}
\usepackage{url}
\usepackage{graphicx}
\usepackage{xspace}

\usepackage{amsmath} %
\usepackage{mathrsfs} %
\usepackage{etoolbox}
\usepackage{cleveref}
\usepackage{tcolorbox}
\usepackage{xcolor}
\usepackage{colortbl}
\usepackage{booktabs}       %
\usepackage{amsfonts}       %
\usepackage{nicefrac}       %
\usepackage{microtype}      %
\usepackage{subcaption}
\usepackage{algorithm}
\usepackage{algorithmic}
\usepackage{multirow}
\usepackage{subcaption}
\usepackage{lipsum}
\usepackage{amssymb}   
\usepackage{xspace}
\usepackage{wrapfig}

\usepackage{pifont} 
\usepackage{xcolor} 
\usepackage{adjustbox} 

\usepackage{array}
\usepackage{adjustbox}
\usepackage{makecell}
\usepackage{minitoc}

\usepackage{enumitem}%
\setlist[itemize]{noitemsep, topsep=0pt}
\usepackage{enumitem,kantlipsum}

\newlength\savewidth

\definecolor{baselinecolor}{HTML}{d6eaf8}

\definecolor{mygray}{gray}{0.4}

\usepackage{xcolor}

\newcommand{\kai}[1]{\textcolor{black}{#1}}

\definecolor{qiucolor}{HTML}{B5651D}
\newcommand{\jqq}[1]{\textcolor{black}{#1}}

\AtBeginEnvironment{tcolorbox}{\tiny}

\ifhtmlout\else
\makeatletter

\patchcmd{\@maketitle}
  {{\LARGE\sc \@title\par}}
  {{\LARGE\bfseries\centering \@title\par}}
  {}
  {}  

\patchcmd{\@maketitle}
  {\begin{tabular}[t]{l}\bf\rule{\z@}{24pt}\@author\end{tabular}}
  {\centering
   \begin{tabular}[t]{>{\small}c}
   \rule{\z@}{24pt}\@author
   \end{tabular}}
  {}
  {}  

\makeatother
\fi

\title{AffineTok: Semantic Affine Consistency for Diffusion-Friendly Visual Tokenizer}

\usepackage{array}
\usepackage{etoolbox}

\ifhtmlout\else
\makeatletter
\patchcmd{\@maketitle}
  {\begin{tabular}[t]{l}\bf\rule{\z@}{24pt}\@author\end{tabular}}
  {\centering
   \begin{tabular}[t]{>{\small}c}
   \rule{\z@}{24pt}\@author
   \end{tabular}}
  {}
\makeatother
\fi

\ifhtmlout
\author{%
Junqiu Yuu\textsuperscript{1,2,*}, Pandeng Li\textsuperscript{2,\dag,\ddag}, Yikai Wang\textsuperscript{1,\dag}, Jiaxing Zhao\textsuperscript{2}, Yujie Wei\textsuperscript{2}, Kaixun Jiang\textsuperscript{2}, Quanhao Li\textsuperscript{2}, Hongtao Yu\textsuperscript{2}, Zhihang Liu\textsuperscript{2}, Zhaohe Liao\textsuperscript{2}, Junjie Zhou\textsuperscript{2}, Yun Zheng\textsuperscript{2}, Yu Liu\textsuperscript{2}, Yanwei Fu\textsuperscript{1,\dag}%
}
\date{%
\textsuperscript{1}Fudan University,\quad
\textsuperscript{2}Alibaba Token Hub, Alibaba Group.\newline
\textsuperscript{*}Work done during an internship at WanTeam, ATH, Alibaba Group.\newline
\textsuperscript{\dag}Corresponding author,\quad
\textsuperscript{\ddag}Project leader.\newline
Emails: yujq24@m.fudan.edu.cn,\quad yanweifu@fudan.edu.cn\newline
Project page: https://michaelyu781.github.io/AffineTok-site/}
\else

\author{%
\bfseries
Junqiu Yu\textsuperscript{1,2}\thanks{Work done during an internship at WanTeam, ATH, Alibaba Group.}~~~
Pandeng Li\textsuperscript{2}\textsuperscript{\dag,\ddag}~~~
Yikai Wang\textsuperscript{1}\textsuperscript{\dag}~~
Jiaxing Zhao\textsuperscript{2}~~
Yujie Wei\textsuperscript{2}~~~
Kaixun Jiang\textsuperscript{2}~~~
Quanhao Li\textsuperscript{2}\\
\bfseries
Hongtao Yu\textsuperscript{2}~~~
Zhihang Liu\textsuperscript{2}~~~
Zhaohe Liao\textsuperscript{2}~~~
Junjie Zhou\textsuperscript{2}~~~
Yun Zheng\textsuperscript{2}~~~
Yu Liu\textsuperscript{2}~~~
Yanwei Fu\textsuperscript{1}\textsuperscript{\dag}
}
\fi

\providecommand{\iclrfinalcopy}{}
\iclrfinalcopy 
\begin{document}


\maketitle
\ifhtmlout\else
\vspace*{-2.66em}
{\centering\small
\textsuperscript{1}Fudan University~~~~~~~~~~
\textsuperscript{2}Alibaba Token Hub, Alibaba Group\\
\textsuperscript{\dag}Corresponding author,\quad
\textsuperscript{\ddag}Project leader\\
\texttt{yujq24@m.fudan.edu.cn}~~~~~~
\texttt{yanweifu@fudan.edu.cn}\\
Project page: \url{https://michaelyu781.github.io/AffineTok-site/}\par}
\vspace*{0.29em}
\fi

\input{sections/abstract}
\input{sections/intro} 
\input{sections/method}
\input{sections/experiments}
\input{sections/related_work}

\input{sections/conclusion}

\bibliography{iclr2026_conference}
\bibliographystyle{iclr2026_conference}



\end{document}

%% file: math_commands.tex
\usepackage{amsmath,amsfonts,bm}

\def\eqref#1{equation~\ref{#1}}

\def\1{\bm{1}}

\DeclareMathAlphabet{\mathsfit}{\encodingdefault}{\sfdefault}{m}{sl}
\SetMathAlphabet{\mathsfit}{bold}{\encodingdefault}{\sfdefault}{bx}{n}

\newcommand{\sacmetric}{\ensuremath{\mathcal{M}_{\mathrm{SAC}}}}
\newcommand{\sacmetricltwo}{\ensuremath{\mathcal{M}_{\mathrm{SAC}_{\ell_2}}}}

%% file: sections/abstract.tex
\begin{figure}[H]
    \centering
    \includegraphics[width=\linewidth]{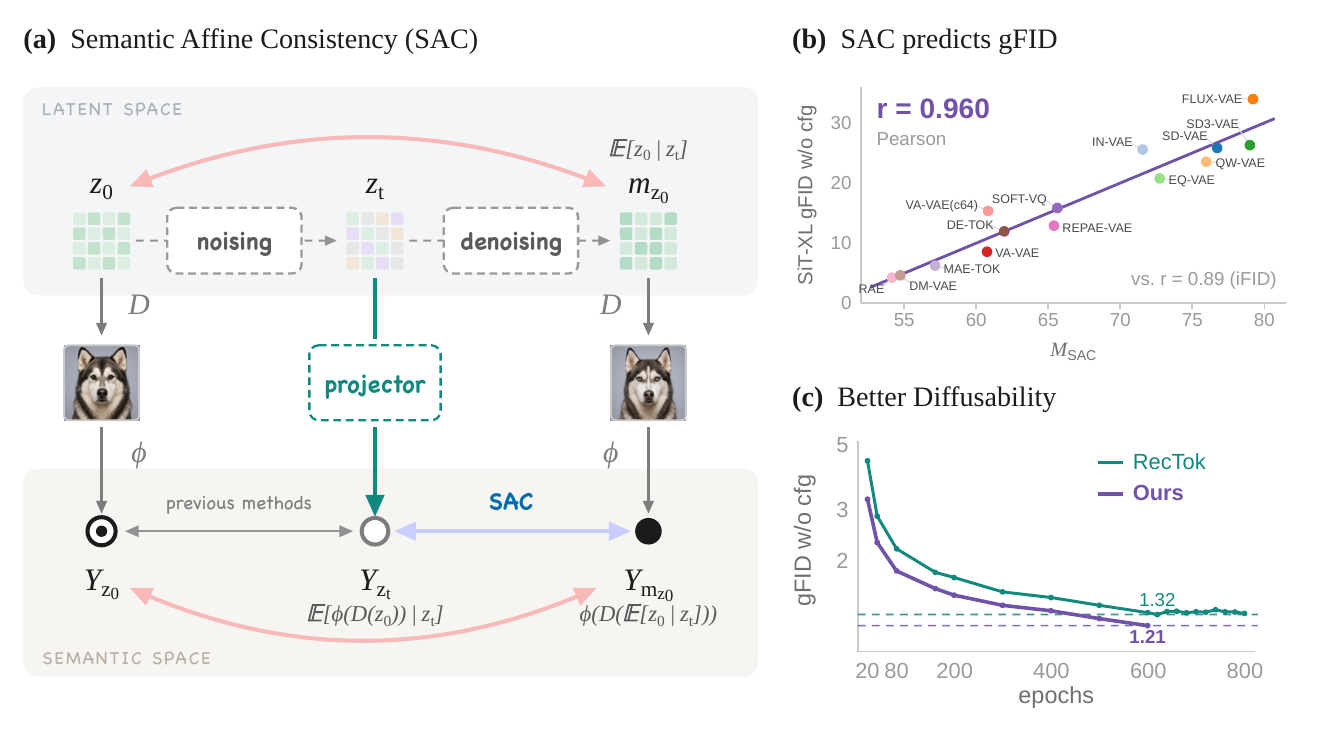}
    \caption{
        \textbf{Semantic Affine Consistency for diffusion-friendly visual tokenizers.}
        \textbf{(a)} Existing methods predict semantics directly from a noisy latent,
        which are the \emph{average of clean-image semantics}. Diffusion instead
        predicts the posterior-mean clean latent and then decodes its semantics, which
        are the \emph{semantics of the averaged clean latent}. SAC requires
        these two predictions to agree, a condition absent from prior supervision.
        \textbf{(b)} Our tokenizer-side metric \sacmetric{} strongly correlates with
        SiT-XL gFID across 14 visual tokenizers, reaching a Pearson correlation of
        \(0.960\).
        \textbf{(c)} SAC-guided training improves diffusability, reaching \(1.21\)
        gFID at 600 epochs without CFG.
    }
    \label{fig:teaser}
    \vspace{-0.8em}
\end{figure}

\begin{abstract}
Visual tokenizers increasingly inject semantic supervision into latent spaces to make downstream diffusion easier. 
Yet how these semantics should be organized to facilitate denoising remains underexplored. 
\kai{
In this paper, we define the semantic recovery objective: the denoising process should recover the semantic content of the ground-truth image from noisy latent, and a good tokenizer should make it easier.
Existing approaches train a projector to predict the semantics directly from the noisy latent.
We argue that this predicts the \emph{average
of clean-image semantics}, whereas what really needs to be aligned is the \emph{semantic
of averaged clean latents}.
More importantly, we demonstrate that the semantic recovery error \emph{orthogonally}
decomposes into the error of the optimal semantic prediction directly from
the noisy latent and the error between these two semantic
predictions. 
}
We therefore identify their consistency as the missing requirement
and call it \emph{Semantic Affine Consistency} (SAC).
To examine whether this overlooked requirement is closely related to
downstream generation, we introduce \sacmetric{}, a tokenizer-side proxy for
SAC. Across the evaluated tokenizers and diffusion model scales,
\sacmetric{} closely tracks generation quality, reaching a
Pearson correlation of $0.960$ with SiT-XL gFID, thereby establishing SAC
as a practical diagnostic and motivating SAC-guided tokenizer training.
We then introduce \emph{AffineTok}, which promotes SAC
through two complementary, training-only components. \emph{Global Semantic Coordination Token} (GSCT)
coordinates the semantic organization of clean latents, keeping semantic
averaging meaningful, while
\emph{Posterior-Mean Semantic Alignment} (PMSA) predicts posterior-mean latents from
noisy inputs and supervises their semantics.
On ImageNet \(256\), compared with the baseline, AffineTok reduces gFID by \(26\%\) at 20 epochs and, with continued training, achieves a new state-of-the-art gFID of \(1.21\)  without classifier-free guidance and 1.10 with guidance.

\end{abstract}

%% file: sections/intro.tex
\section{\jqq{Introduction}}
\label{sec:intro}

In latent diffusion \kai{models}, a visual tokenizer is more than a compressor: 
it determines the geometry of the denoising problem that the
model must learn. This realization has driven a recent shift from tokenizers
optimized primarily for reconstruction toward latent spaces explicitly shaped
for generation. In particular, modern approaches inject features from visual
foundation models (VFMs) into clean latents~\citep{yao2025reconstruction},
regularize their structure~\citep{kouzelis2025eq,liu2025delving}, or extend
semantic supervision to noisy latent states~\citep{shi2025rectok,page2026boosting}.
These methods share an appealing premise: if semantics are more accessible in
the latent space, diffusion should be easier. 
Yet in diffusion, the semantic outcome is
determined by decoding the predicted clean latent, rather than by directly
reading the noisy state.
This distinction raises a central
question: \emph{how should a tokenizer organize semantics in the latent space to make
diffusion easier to learn?}

\jqq{To answer this question, our starting point is the goal of denoising: to recover
the clean image from a noisy observation, which involves recovering not only
its latent representation but also its semantic content.} Achieving this goal requires aligning 
clean-image semantics with those decoded from the predicted clean latent, 
with the latter corresponding to the \emph{semantics of the averaged clean latent}.
As shown in Fig.~\ref{fig:teaser}(a), existing methods instead apply semantic
supervision through a different path, training a semantic projector to predict
the clean-image semantics directly from the noisy latent. Under squared loss, 
the optimal projector predicts the \emph{average of clean-image semantics}.
As semantic mapping is nonlinear, averaging and semantic mapping need not commute,
so current supervision does not ensure the required
alignment of the decoded semantics. In fact, it can be exactly proven the semantic error of the
image decoded from diffusion's clean-latent prediction decomposes
orthogonally into two terms: the error of predicting semantics directly from the noisy
latent and the gap between the average of semantics and the semantics of the averaged latent.
Closing this gap requires the two semantic predictions to be consistent. We call this
agreement \emph{Semantic Affine Consistency} (SAC).

\jqq{
After identifying SAC as a requirement overlooked by existing
tokenizer methods, we next ask whether tokenizers that better satisfy SAC
support stronger downstream generation.
}
Directly evaluating SAC would require training a downstream diffusion model for each tokenizer. 
Inspired by iFID~\citep{xu2026making}, we
instead construct probe points by interpolating between neighboring clean
latents, using them as local proxies for posterior averaging. The resulting
tokenizer-side probe metric, \sacmetric{}, measures SAC violations at these
points.
Across the 14 visual tokenizers and diffusion model scales,
\sacmetric{} closely tracks downstream generation quality (Fig.~\ref{fig:teaser}(b)),
reaching a Pearson correlation of $0.960$ with SiT-XL gFID.

SAC also guides tokenizer learning. Rather than enforcing SAC only at these
probe points, which provide limited coverage of the latent space, we bring
posterior-mean semantics under supervision through two complementary,
training-only components. The
\emph{Global Semantic Coordination Token} (GSCT) uses a VFM-supervised global
token to coordinate the organization of clean patch latents across images,
keeping cross-image semantic averaging meaningful. The
\emph{Posterior-Mean Semantic Alignment} (PMSA) uses a lightweight
predictor to estimate posterior-mean latents from noisy inputs and a semantic
projector calibrated on clean latents to supervise their semantics.
Both components are removed after tokenizer training, leaving the
downstream diffusion training unchanged. As shown in
Fig.~\ref{fig:teaser}(c), on ImageNet \(256\times256\), the resulting tokenizer improves gFID throughout
diffusion training and ultimately reaches $1.21$ without classifier-free guidance (cfg) and $1.10$
with cfg.

Our contributions are as follows:
\begin{itemize}
    \item We identify a previously overlooked mismatch between the semantics
    directly predicted from a noisy latent and those decoded from its
    posterior-mean clean-latent prediction. We formalize their agreement as SAC
    and derive an exact orthogonal decomposition showing their
    disagreement corresponds to a previously unaddressed term of semantic recovery error.
    
    \item We introduce the tokenizer-side metric \sacmetric{}, a proxy for SAC based
    on interpolation between neighboring clean latents. By measuring both the
    coefficient and direction of decoded semantic transitions, it closely
    tracks generation quality across tokenizers and diffusion model scales
    without training downstream diffusion models.

    \item We develop a SAC-guided tokenizer-training method with GSCT and PMSA.
    GSCT coordinates the organization of clean patch latents across images,
    keeping cross-image semantic averaging meaningful, while PMSA predicts
    posterior-mean latents from noisy inputs and supervises their semantics.
    Both components are tokenizer training-only, 
    preserving the downstream diffusion training unchanged.
\end{itemize}

%% file: sections/method.tex
\section{Method}
\label{sec:method}

\begin{figure}[t]
    \centering
    \includegraphics[width=\linewidth]{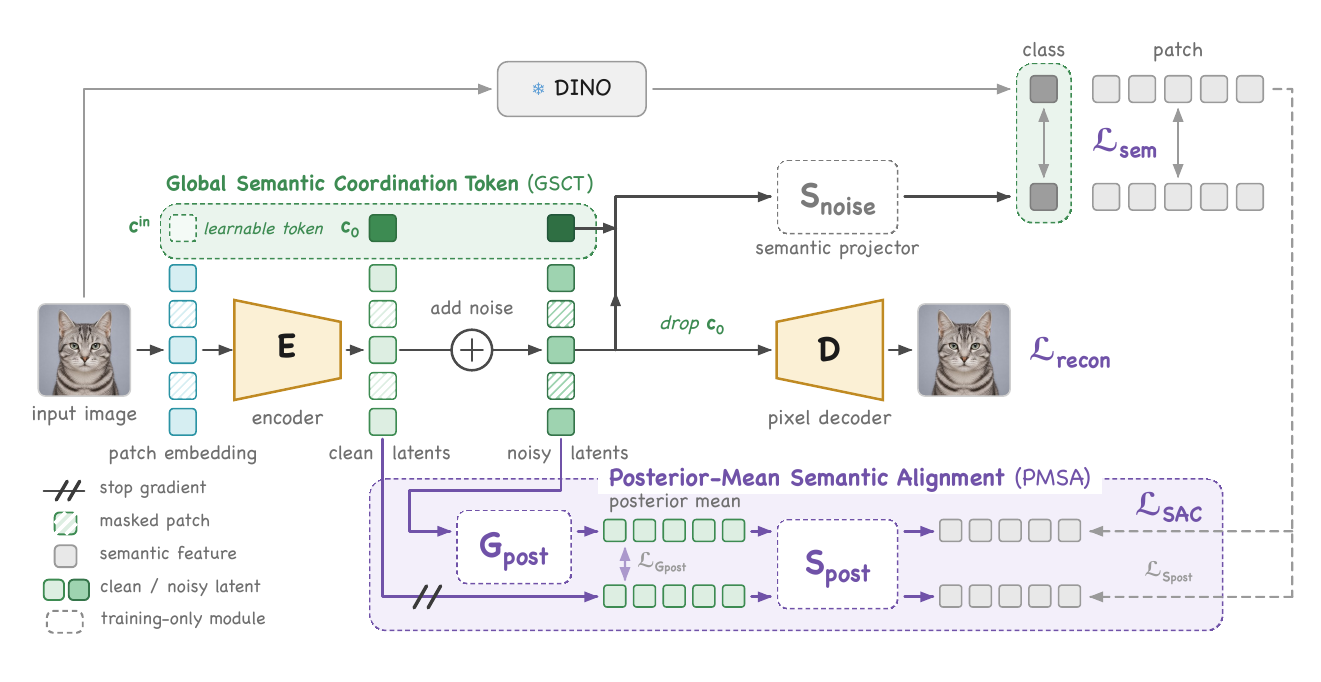}
    \caption{
        \textbf{Overall framework of AffineTok.}
        The input image is processed in parallel by a frozen VFM and a trainable
        tokenizer. Its encoder jointly maps patch embeddings and a prepended
        learnable GSCT to a global output and clean patch latents \(z_0\). After \(z_0\) is noised
        into \(z_t\), the resulting global and noisy patch representations receive
        semantic supervision; the global output is then discarded, while \(z_t\) is
        decoded for reconstruction. Along the PMSA path, \(G_{\mathrm{post}}\) maps
        \(z_t\) to a posterior-mean estimate, and \(S_{\mathrm{post}}\) maps the
        estimate to semantics. Only clean patch latents are retained downstream.
    }
    \label{fig:sac-overview}
\end{figure}

\jqq{
We study how semantic supervision in visual tokenizers supports semantic
recovery required by downstream diffusion. In
Sec.~\ref{sec:semantic_affine_consistency}, we decompose the semantic recovery
error into the semantic uncertainty in a noisy latent and the gap
between the semantics directly predicted from the noisy latent and those decoded from
the posterior-mean clean latent. We define their agreement as
\emph{Semantic Affine Consistency} (SAC). In
Sec.~\ref{sec:sac_metric}, we introduce the tokenizer-side metric
\sacmetric{}, a proxy for SAC based on interpolation between neighboring clean
latents. In Sec.~\ref{sec:tokenizer_training}, we use
this analysis to guide tokenizer training by extending semantic supervision to
previously unconstrained posterior-mean estimates. The training framework
is shown in Fig.~\ref{fig:sac-overview}.
}

\vspace{-1.3em}

\subsection{\jqq{Semantic Affine Consistency}}
\label{sec:semantic_affine_consistency}

The goal of denoising is to recover the clean image from a noisy observation.
In latent diffusion, this requires recovering not only the clean latent but
also the semantic content of the image obtained by decoding it. We refer to
the latter objective as \emph{semantic recovery}.

Specifically, given an image \(X\sim p_{\mathrm{data}}\), we define
\begin{equation}
    z_0=E(X),\qquad
    Y_X=\phi(X),\qquad
    z_t=(1-t)z_0+t\epsilon,
    \quad \epsilon\sim\mathcal{N}(0,I),
    \label{eq:sac_setup}
\end{equation}
where \(E\) and \(D\) are the tokenizer's encoder and decoder,
and \(\phi\) is a frozen visual foundation model.

Under squared denoising loss, the Bayes-optimal clean-latent prediction
from \(z_t\) is the posterior mean
\begin{equation}
    m_{z_0}(z_t,t)
    :=
    \mathbb{E}[z_0\mid z_t,t].
    \label{eq:sac_latent_posterior_mean}
\end{equation}
Decoding this prediction and applying \(\phi\) gives the corresponding
semantic prediction
\begin{equation}
    Y_{m_{z_0}}(z_t,t)
    :=
    \phi\circ D\!\left(m_{z_0}(z_t,t)\right).
    \label{eq:sac_latent_path_prediction}
\end{equation}
Therefore, the error of the \emph{semantic recovery} objective at
\((z_t,t)\) under squared semantic loss is
\begin{equation}
    \mathcal{E}_{\mathrm{sem}}(z_t,t)
    :=
    \mathbb{E}\!\left[
        \left\|
            Y_X-Y_{m_{z_0}}(z_t,t)
        \right\|_2^2
        \,\middle|\, z_t,t
    \right].
    \label{eq:downstream_semantic_error}
\end{equation}

To further analyze this error, we first separate the component caused by
insufficient semantic information contained in \(z_t\) after noising. To do
so, we consider the theoretically optimal semantic prediction given \((z_t,t)\). Under
squared semantic loss, this prediction is the conditional mean
\begin{equation}
    Y_{z_t}(z_t,t)
    :=
    \arg\min_{y}
    \mathbb{E}\!\left[
        \|Y_X-y\|_2^2
        \,\middle|\,z_t,t
    \right]
    =
    \mathbb{E}[Y_X\mid z_t,t].
    \label{eq:sac_optimal_semantic_prediction}
\end{equation}

Further, it can be proved that the semantic recovery error admits the
following exact decomposition:
\begin{equation}
\begin{aligned}
    \mathcal{E}_{\mathrm{sem}}(z_t,t)
    &=
    \mathbb{E}\!\left[
        \left\|
            Y_X-Y_{z_t}(z_t,t)
        \right\|_2^2
        \,\middle|\,z_t,t
    \right]
    +
    \left\|
        Y_{z_t}(z_t,t)
        -
        Y_{m_{z_0}}(z_t,t)
    \right\|_2^2.
\end{aligned}
\label{eq:sac_semantic_error}
\end{equation}
The full derivation is provided in
Appendix~\ref{app:sac_derivation}.
Previous methods introduce a projector \(P\) to predict the clean-image
semantics from noisy latents~\citep{shi2025rectok,page2026boosting}, using:
\begin{equation}
  \mathcal{L}_{\mathrm{noise}}
  =
      \left\|Y_X - P(z_t,t)\right\|_2^2.
  \label{eq:noisy_semantic_regression}
\end{equation}
For a fixed encoder \(E\), the optimal projector satisfies
$
  P^\star(z_t,t)
  =
  \mathbb{E}[Y_X\mid z_t,t]
  =
  Y_{z_t}(z_t,t).
$
Substituting \(P^\star\) into
Eq.~\ref{eq:noisy_semantic_regression} shows that its minimum is exactly the
expectation of the first term in Eq.~\ref{eq:sac_semantic_error}. Therefore,
the loss encourages the encoder to reduce the first term.

However, the second term in Eq.~\ref{eq:sac_semantic_error} reveals a missing
requirement. In fact, \(Y_{z_t}(z_t,t)\) is the \emph{average of clean-image semantics},
whereas \(Y_{m_{z_0}}(z_t,t)\) is the \emph{semantic of an averaged clean
latent}. As the semantic decoding map \(g=\phi\circ D\) is generally nonlinear,
semantic decoding and averaging need not commute, so the two need not agree.
We call this missing requirement \emph{Semantic Affine Consistency} (SAC). A
tokenizer satisfies SAC when
\(Y_{m_{z_0}}(z_t,t)=Y_{z_t}(z_t,t)\), namely,
\begin{equation}
    \phi\circ D\!\left(\mathbb{E}[z_0\mid z_t,t]\right)
    =
    \mathbb{E}[\phi\circ D(z_0)\mid z_t,t].
    \label{eq:semantic_affine_condition}
\end{equation}


\paragraph{Derivation of Eq.~\ref{eq:sac_semantic_error}.}
\label{app:sac_derivation}
For brevity, let \(Y_{z_t}=Y_{z_t}(z_t,t)\) and
\(Y_{m_{z_0}}=Y_{m_{z_0}}(z_t,t)\). Both are fixed when conditioning on
\(z_t\) and \(t\). Expanding
\(
Y_X-Y_{m_{z_0}}
=
(Y_X-Y_{z_t})+(Y_{z_t}-Y_{m_{z_0}})
\)
gives the two terms in Eq.~\ref{eq:sac_semantic_error} and the cross term
\begin{equation}
\begin{aligned}
    &\mathbb{E}\!\left[
        (Y_X-Y_{z_t})^\top(Y_{z_t}-Y_{m_{z_0}})
        \mid z_t,t
    \right]
    =
    \left(
        \mathbb{E}[Y_X\mid z_t,t]-Y_{z_t}
    \right)^\top
    (Y_{z_t}-Y_{m_{z_0}})
    =0.
\end{aligned}
\end{equation}
Therefore, the decomposition holds exactly.

\begin{figure}[t]
    \centering
    \includegraphics[width=\linewidth]{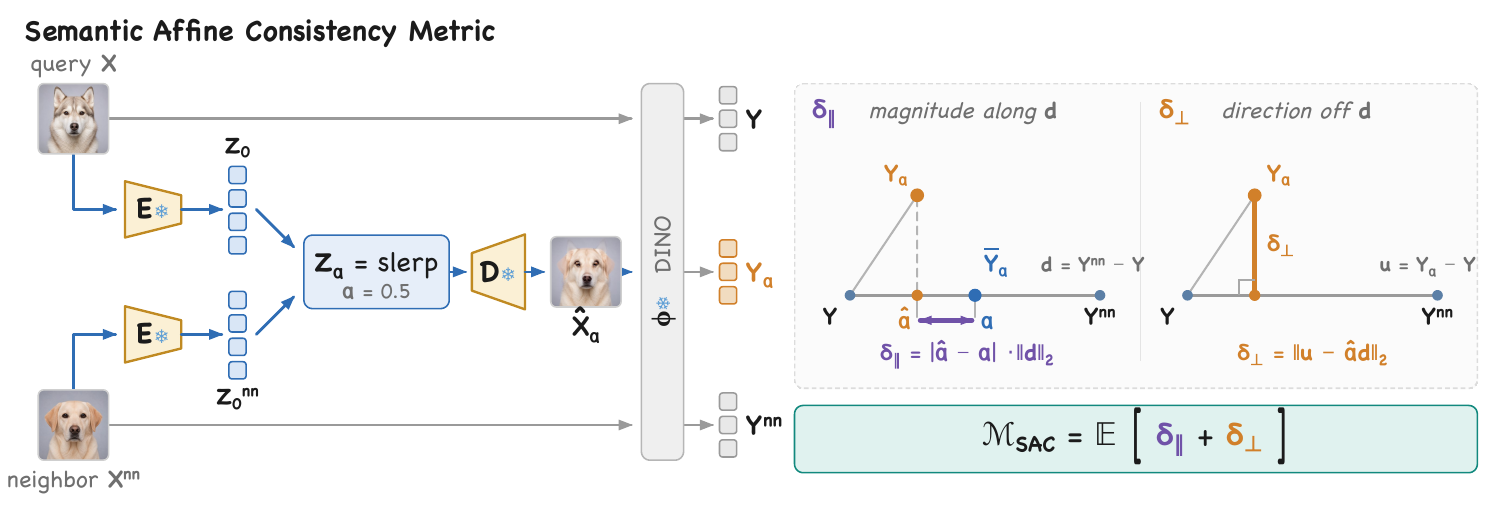}
    \caption{
        \textbf{SAC probing.}
        We compute \sacmetric{} using neighboring latent interpolation to measure
        the agreement between the average of semantics and the semantics
        decoded from the averaged latent.
    }
    \label{fig:sic}
\end{figure}

\subsection{
\jqq{SAC Aligns with Downstream Generation Capacity}
}
\label{sec:sac_metric}
A natural question is how SAC, a requirement overlooked by existing methods,
relates to downstream generation: do tokenizers that better satisfy SAC support
stronger downstream generation?

In practice, directly evaluating
Eq.~\ref{eq:semantic_affine_condition} would require estimating both $Y_{z_t}(z_t,t)$ and $Y_{m_{z_0}}(z_t,t)$
over the full diffusion posterior. 
Inspired by~\citep{xu2026making}, we approximate local posterior averaging
with a two-point support formed by a clean latent \(z_0\) and its nearest
neighbor \(z_0^{\mathrm{nn}}\). Their interpolated latent \(z_\alpha\) serves
as a proxy for the posterior mean \(m_{z_0}\). Correspondingly, the semantics
decoded from \(z_\alpha\) approximate \(Y_{m_{z_0}}\), while the interpolation
of their clean-image semantics approximates \(Y_{z_t}\). We therefore probe SAC
by comparing these two semantic quantities.

Specifically, let \(z_0=E(X)\) be a query latent and
\(z_0^{\mathrm{nn}}=E(X^{\mathrm{nn}})\) its nearest neighbor in a latent
bank. We fix \(\alpha=0.5\) and form
\begin{equation}
    z_\alpha
    =
    \operatorname{slerp}
    \left(
        z_0,z_0^{\mathrm{nn}};\alpha
    \right),
    \qquad
    \hat{X}_\alpha
    =
    D(z_\alpha).
    \label{eq:sac_metric_interpolation}
\end{equation}
We extract the endpoint semantics and the semantics decoded from the
interpolated latent as
\begin{equation}
    Y=\phi(X),
    \qquad
    Y^{\mathrm{nn}}=\phi(X^{\mathrm{nn}}),
    \qquad
    Y_\alpha=\phi(\hat{X}_\alpha).
    \label{eq:sac_metric_features}
\end{equation}

Let \(d\) and \(u\) denote the semantic differences from \(Y\) to
\(Y^{\mathrm{nn}}\) and \(Y_\alpha\), respectively:
\begin{equation}
\begin{aligned}
    d
    &=
    Y^{\mathrm{nn}}-Y,
    \qquad
    u
    =
    Y_\alpha-Y,
    \qquad
    \hat{\alpha}
    =
    \frac{\langle u,d\rangle}{\|d\|_2^2}.
\end{aligned}
\label{eq:sac_metric_projection}
\end{equation}
The target semantic difference is \(\alpha d\). To measure the error in
interpolation coefficient and semantic direction, we further decompose the difference
between \(u\) and \(\alpha d\) into components parallel and orthogonal to
\(d\):
\begin{equation}
\begin{aligned}
    \delta_{\parallel}
    &=
    |\hat{\alpha}-\alpha| \cdot \,\|d\|_2,
    \qquad
    \delta_{\perp}
    =
    \|u-\hat{\alpha}d\|_2.
\end{aligned}
\label{eq:sac_metric_target}
\end{equation}

As shown in Fig.~\ref{fig:sic}, 
we define the tokenizer-side SAC metric \sacmetric{} as
\begin{equation}
    \sacmetric
    =
    \mathbb{E}_{X,X^{\mathrm{nn}}}
    \left[
        \delta_{\parallel}
        +
        \delta_{\perp}
    \right].
    \label{eq:sac_metric}
\end{equation}
A lower \sacmetric{} indicates that latent interpolation more closely
preserves the corresponding semantic interpolation. As shown in
Fig.~\ref{fig:teaser}(b), across 14 visual tokenizers,
\sacmetric{} achieves a Pearson
correlation of \(0.960\) with SiT-XL gFID.

\subsection{SAC-Guided Tokenizer Training}
\label{sec:tokenizer_training}

Having established \sacmetric{} as a tokenizer-side probe of SAC, we next
ask how SAC can guide tokenizer training. A straightforward approach would be
to directly optimize Eq.~\ref{eq:sac_metric} over sampled interpolation
pairs. However, enforcing consistency over this restricted set of latent-space
locations does not ensure SAC throughout the latent space.

Before introducing our SAC-guided components, we specify the base
objective for tokenizer training:
\begin{equation}
    \mathcal{L}_{\mathrm{tok}}
    =
    \mathcal{L}_{\mathrm{rec}}
    +
    \lambda_{\mathrm{sem}}\mathcal{L}_{\mathrm{sem}},
    \label{eq:base_tokenizer_objective}
\end{equation}
where \(\mathcal{L}_{\mathrm{rec}}\) aggregates pixel-wise reconstruction loss,
LPIPS loss, GAN loss~\citep{van2017neural}, and KL loss~\citep{kingma2013auto},
while \(\mathcal{L}_{\mathrm{sem}}\)~\citep{shi2026rectok} combines masked feature reconstruction with
noisy-latent alignment to frozen VFM features, thereby injecting clean-image
semantics into noisy latents and reducing the semantic uncertainty captured by
the first term of Eq.~\ref{eq:sac_semantic_error}.
While this constrains the direct noisy-latent semantic path
\(Y_{z_t}(z_t,t)\), it leaves \(Y_{m_{z_0}}(z_t,t)\), the semantics recovered
through diffusion's clean-latent prediction, outside the training objective.

Completing semantic recovery and satisfying SAC require bringing this
missing posterior-mean path under supervision.
We therefore introduce two complementary, training-only components.
GSCT (Sec.~\ref{sec:gsct}) promotes a consistent organization of clean-latent
semantics, keeping cross-image semantic averaging meaningful, while PMSA
(Sec.~\ref{sec:pmsa}) predicts posterior-mean latents
from noisy inputs and explicitly supervises their semantics.

\subsubsection{Global Semantic Coordination Token}
\label{sec:gsct}
To make latents more semantic, previous methods align them with VFM
features~\citep{yao2025reconstruction,page2026boosting}, while
RecTok~\citep{shi2025rectok} further combines such alignment with masking to
encourage patch latents to capture more global semantic information. While
effective, masking does not explicitly coordinate how this global information
is incorporated into the patch latents, so semantic organization may
become less consistent across images, as shown in
Fig.~\ref{fig:semantic-probes}. We therefore introduce a
\emph{Global Semantic Coordination Token} (GSCT) to provide such global
coordination.

GSCT complements position-wise semantic alignment with a joint constraint on
the global information carried by all patch latents. Specifically, let
\(P=[p^1,\ldots,p^N]\) denote the unmasked patch-token sequence supplied to the
encoder. We prepend a learned token \(c^{\mathrm{in}}\) and obtain
\begin{equation}
    [\,c_0;z_0\,]
    =
    E([\,c^{\mathrm{in}};P\,]),
    \label{eq:gsct_encoding}
\end{equation}
where \(c_0\) is the GSCT output and \(z_0\) contains the patch latents.

A training-only semantic projector reads the full encoder output as
\([\,\hat{y}^{\mathrm{gsct}};\hat{Y}^{\mathrm{patch}}\,]
=S_{\mathrm{noise}}([\,c_0;z_0\,])\).
Given the frozen VFM features
\(\phi(X)=[\,y^{\mathrm{cls}};Y^{\mathrm{patch}}\,]\), the patch predictions
retain their original semantic supervision, while the GSCT prediction is
aligned with the global VFM feature, (e.g. DINO's class token):
\begin{equation}
    \mathcal{L}_{\mathrm{GSCT}}
    =
    d_{\cos}\!\left(
        \hat{y}^{\mathrm{gsct}},
        \operatorname{sg}[y^{\mathrm{cls}}]
    \right).
    \label{eq:gsct_global_loss}
\end{equation}

Because \(c_0\) exchanges information with all patch tokens during encoding,
its alignment with the global VFM feature jointly guides how patch latents
capture global information under masking. This shared semantic reference
coordinates the semantic organization of clean latents across images, 
keeping cross-image semantic averaging meaningful.
The GSCT output is discarded before pixel decoding, leaving the
downstream latent interface unchanged.

\begin{figure}[t]
    \centering
    \includegraphics[width=\linewidth]{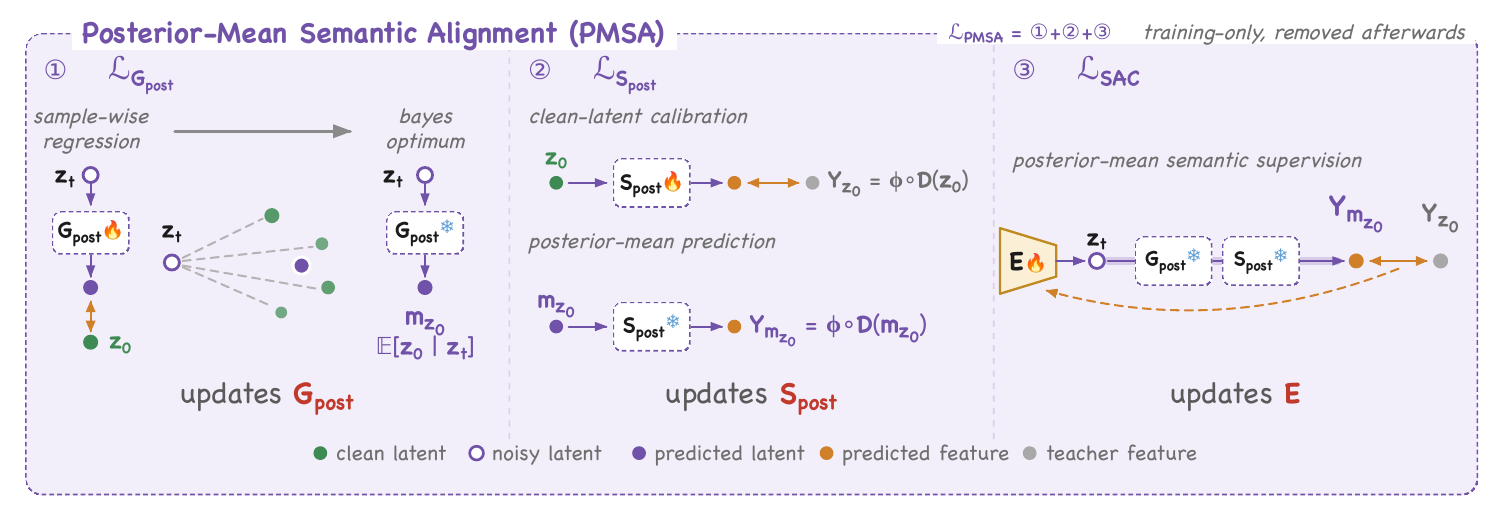}
    \caption{
        \textbf{PMSA training.}
        PMSA learns a posterior-mean predictor, calibrates a
        semantic projector on clean latents, and supervises the semantics recovered
        from predicted clean latents.
    }
    \label{fig:pmsa-training}
\end{figure}

\subsubsection{Posterior-Mean Semantic Alignment}
\label{sec:pmsa}

So far, we have injected semantic information into \(z_t\) through
\(\mathcal{L}_{\mathrm{sem}}\) and further ensured consistent semantic
coordinates across images through \(\mathcal{L}_{\mathrm{GSCT}}\).
However, as shown in Sec.~\ref{sec:semantic_affine_consistency}, we still
need to explicitly constrain the semantics decoded from the clean-latent
prediction, namely, \(Y_{m_{z_0}}(z_t,t)\).

In fact, directly constraining \(Y_{m_{z_0}}\) raises two practical issues:
(1) how to estimate the posterior mean from a noisy latent; and
(2) how to avoid the heavy computation through the pixel decoder and frozen
VFM at every training step. To address these issues, we introduce two
training-only modules: a posterior predictor \(G_{\mathrm{post}}\) and a
semantic projector \(S_{\mathrm{post}}\). The former estimates the posterior
mean from a noisy latent, while the latter approximates the decoded VFM
semantics directly in latent space.
The three PMSA training objectives are summarized in
Fig.~\ref{fig:pmsa-training}.

\textbf{Estimate the posterior mean.}
Our first target is to estimate the posterior mean
\(m_{z_0}(z_t,t)\) required by
Eq.~\ref{eq:semantic_affine_condition} from the noisy latent \(z_t\).
We follow the principle used in diffusion models:
\emph{supervision from individual clean samples is sufficient to estimate the
posterior mean from a noisy input}~\citep{karras2022elucidating}.
Specifically, although each \(z_t\) is paired with only one clean target
\(z_0\), the population optimum of squared regression is exactly the
posterior mean, namely,
\begin{equation}
    G_{\mathrm{post}}^{*}(z_t,t)
    =
    \mathbb{E}[z_0\mid z_t,t]
    =
    m_{z_0}(z_t,t).
    \label{eq:pmsa_posterior_optimum}
\end{equation}
We therefore train \(G_{\mathrm{post}}\) with
\begin{equation}
    \mathcal{L}_{G_{\mathrm{post}}}
    =
    \lambda_z
    \left\|
        G_{\mathrm{post}}(z_t,t)-z_0
    \right\|^2.
    \label{eq:pmsa_mean_loss}
\end{equation}

\textbf{Calibrating the semantic projector.}
Our second target is to approximate the decoded VFM semantics
\(\phi\circ D(Z)\) directly in latent space, avoiding repeated computation
through the pixel decoder and frozen VFM. We learn a semantic projector
\(S_{\mathrm{post}}\) for this purpose. Given a latent sequence \(z\),
it predicts one global feature and \(N\) patch features:
\begin{equation}
    \widehat{Y}(z)
    =
    S_{\mathrm{post}}(z)
    =
    [\,\hat{y}^{0};\hat{y}^{1},\ldots,\hat{y}^{N}\,],
    \qquad
    Y_X
    =
    [\,y^{0};y^{1},\ldots,y^{N}\,],
    \qquad
    y^{0}=y^{\mathrm{cls}}.
    \label{eq:pmsa_semantic_prediction}
\end{equation}
We measure their semantic distance and calibrate
\(S_{\mathrm{post}}\) on observed clean latents using
\begin{equation}
\begin{aligned}
    \mathcal{D}_{\mathrm{sem}}
    \left(
        \widehat{Y},Y_X
    \right)
    &=
    d_{\cos}(\hat{y}^{0},y^{0})
    +
    \frac{1}{N}
    \sum_{p=1}^{N}
    d_{\cos}(\hat{y}^{p},y^{p}),
    \qquad
    \mathcal{L}_{S_{\mathrm{post}}}
    =
    \lambda_y
    \mathcal{D}_{\mathrm{sem}}\!\left(
        S_{\mathrm{post}}(z_0),Y_X
    \right).
\end{aligned}
\label{eq:pmsa_spost_loss}
\end{equation}
The calibrated semantic projector is then used to approximate
\(\phi\circ D\) on posterior-mean estimates.

\textbf{Semantic supervision for posterior-mean estimates.}
With \(G_{\mathrm{post}}\) estimating \(m_{z_0}(z_t,t)\) and
\(S_{\mathrm{post}}\) approximating its decoded VFM semantics, their
composition
\begin{equation}
    \widehat{Y}_{m_{z_0}}(z_t,t)
    :=
    S_{\mathrm{post}}\!\left(
        G_{\mathrm{post}}(z_t,t)
    \right)
    \label{eq:pmsa_semantic_estimate}
\end{equation}
provides a latent-space approximation to \(Y_{m_{z_0}}(z_t,t)\).

Once \(z_t\) retains sufficient semantic information, as encouraged by
\(\mathcal{L}_{\mathrm{sem}}\), directly supervising
\(\widehat{Y}_{m_{z_0}}(z_t,t)\) with the clean semantics \(Y_{z_0}\) drives
its optimum toward \(Y_{z_t}(z_t,t)\). We therefore use \(Y_{z_0}\) as the
supervision target for simplicity
\begin{equation}
    \mathcal{L}_{\mathrm{SAC}}
    =
    \lambda_y
    \mathcal{D}_{\mathrm{sem}}\!\left(
        S_{\mathrm{post}}\!\left(
            G_{\mathrm{post}}(z_t,t)
        \right),
        Y_X
    \right).
    \label{eq:pmsa_sac_loss}
\end{equation}
By directly supervising the previously unconstrained
\(Y_{m_{z_0}}\), this objective promotes SAC and improves semantic
recovery. We retain \(\mathcal{L}_{\mathrm{sem}}\) throughout training because its direct
semantic supervision on \(z_t\) is easier to learn, thereby reducing the
optimization difficulty of this additional constraint.

\textbf{Overall PMSA objective.}
The complete PMSA objective combines the three losses:
\begin{equation}
    \mathcal{L}_{\mathrm{PMSA}}
    =
    \mathcal{L}_{G_{\mathrm{post}}}
    +
    \mathcal{L}_{S_{\mathrm{post}}}
    +
    \mathcal{L}_{\mathrm{SAC}}.
    \label{eq:pmsa_objective}
\end{equation}

Within PMSA, the three objectives update different parameter sets:
\begin{equation}
\begin{aligned}
    \Delta\theta_G
    &=
    -\eta_G
    \nabla_{\theta_{G_{\mathrm{post}}}}
    \mathcal{L}_{G_{\mathrm{post}}},
    \Delta\theta_S
    =
    -\eta_S
    \nabla_{\theta_{S_\mathrm{post}}}
    \mathcal{L}_{S_{\mathrm{post}}},
    \Delta\theta_E
    =
    -\eta_E
    \nabla_{\theta_E}
    \mathcal{L}_{\mathrm{SAC}},
\end{aligned}
\label{eq:pmsa_parameter_updates}
\end{equation}
where the remaining parameter sets are held fixed in each update.

\subsubsection{Overall Objective}
\label{sec:overall_objective}

The complete tokenizer training objective is
\begin{equation}
    \mathcal{L}_{\mathrm{total}}
    =
    \mathcal{L}_{\mathrm{tok}}
    +
    \lambda_{\mathrm{GSCT}}\mathcal{L}_{\mathrm{GSCT}}
    +
    \lambda_{\mathrm{PMSA}}\mathcal{L}_{\mathrm{PMSA}}.
    \label{eq:overall_training_objective}
\end{equation}

%% file: sections/experiments.tex
\section{Experiments}
\label{sec:experiments}

We first describe the experimental setup in
Sec.~\ref{sec:experimental_setup}.
Then, we evaluate whether \sacmetric{} predicts downstream generation
performance in Sec.~\ref{sec:metric_evaluation}, examine whether GSCT and PMSA
improve tokenizer diffusability in Sec.~\ref{sec:tokenizer_diffusability}, and
present controlled ablations in Sec.~\ref{sec:ablation}.

\subsection{Experimental Setup}
\label{sec:experimental_setup}
\paragraph{Metric evaluation.}
Following~\citet{xu2026making}, we evaluate 14 publicly available visual
tokenizer checkpoints spanning different latent dimensions, architectures, and
training objectives.
We use the SiT-B and SiT-XL gFID results reported in that work to measure
downstream generation quality.
All tokenizer-side metrics are computed on the ImageNet validation split.
We report Pearson, Spearman, and Kendall correlations between each
tokenizer-side metric and gFID.

\paragraph{Tokenizer training.}
We follow the tokenizer architecture and training protocol of
RecTok~\citep{shi2026rectok}, and use DINOv3-S\citep{simeoni2025dinov3} as the VFM.
With a \(16\times\) spatial compression ratio, the tokenizer represents each
image using \(d=128\) latent dimension.
We set \(\lambda_{\mathrm{GSCT}}=\lambda_{\mathrm{PMSA}}=1\).
The posterior predictor \(G_{\mathrm{post}}\) is a four-block
transformer with hidden dimension \(768\), projecting the
noisy latent sequence from \(d\) to \(768\) dimensions and back to \(d\).
The semantic projector \(S_{\mathrm{post}}\) adopts a two-layer transformer
architecture with hidden dimension \(384\).
GSCT and PMSA are used only during tokenizer training, and the exported
tokenizer retains the same latent interface as RecTok.

\paragraph{Diffusion training and evaluation.}
For downstream diffusion training, we use the same
\(\mathrm{DiT}^{\mathrm{DH}}\)-XL~\citep{zheng2026diffusion} backbone and training protocol as RecTok.
All experiments are conducted on ImageNet-1K~\citep{russakovsky2015imagenet} at \(256\times256\) resolution.
We report Fr\'echet Inception Distance (gFID)~\citep{heusel2017gans}, Inception Score (IS)~\citep{salimans2016improved}, precision (Prec.), and recall (Rec.)~\citep{kynkaanniemi2019improved}, computed from 50K
generated samples following the RecTok evaluation protocol.

\begin{figure}[t]
  \centering
  \includegraphics[width=\linewidth]{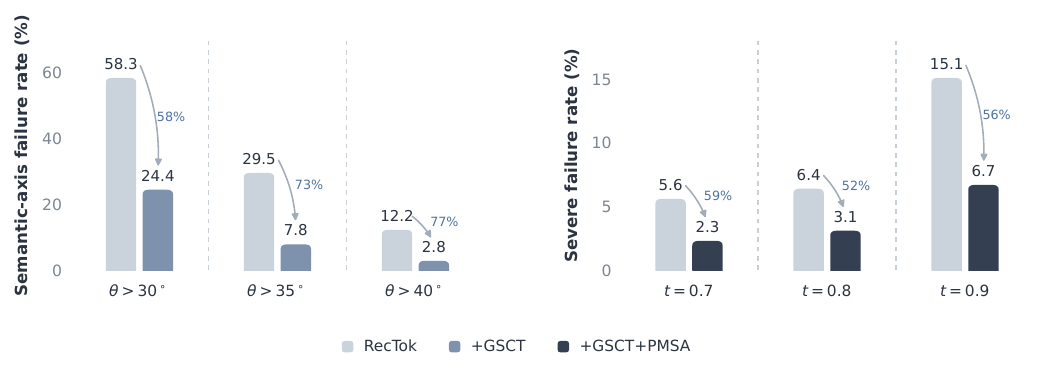}
    \caption{
        \textbf{Targeted semantic probes for GSCT and PMSA (lower is better).}
        \emph{Left:} Cross-split semantic-axis inconsistency rates for RecTok and
        +GSCT across three angular thresholds.
        \emph{Right:} Top-10 semantic retrieval miss rates for RecTok and
        +GSCT+PMSA under three severe noise levels.
    }
   \label{fig:semantic-probes}
\end{figure}

\subsection{The SAC Metric Predicts Downstream Generation}
\label{sec:metric_evaluation}

\begin{table}[t]
\caption{
\textbf{Correlation between tokenizer-side metrics and downstream gFID.}
We report Pearson, Spearman, and Kendall correlations across 14 tokenizers for
SiT-XL and 13 tokenizers for SiT-B.
}
\label{tab:metric_corr}
\centering
\resizebox{\linewidth}{!}{
\begin{tabular}{lcccccc}
\toprule
\multirow{2}{*}{\textbf{Tokenizer-side metric}}
& \multicolumn{3}{c}{\textbf{SiT-XL}}
& \multicolumn{3}{c}{\textbf{SiT-B}} \\
\cmidrule(lr){2-4}\cmidrule(lr){5-7}
& \textbf{Pearson} & \textbf{Spearman} & \textbf{Kendall}
& \textbf{Pearson} & \textbf{Spearman} & \textbf{Kendall} \\
\midrule
  iFID
  & 0.896 & 0.881 & 0.758
  & 0.856 & 0.830 & 0.692 \\

  \sacmetricltwo{}
  & 0.933 & 0.943 & 0.846
  & 0.896 & 0.874 & 0.744 \\
  
  \sacmetric{}
  & \textbf{0.960} & \textbf{0.974} & \textbf{0.912}
  & \textbf{0.946} & \textbf{0.929} & \textbf{0.821} \\
  \bottomrule
\end{tabular}
}
\end{table}

\paragraph{The SAC metric consistently predicts tokenizer diffusability.}
As shown in Table~\ref{tab:metric_corr}, \sacmetric{} achieves the strongest
correlation with downstream gFID for both SiT-B and SiT-XL under every
correlation measure, with its Pearson correlation reaching \(0.960\) for
SiT-XL.
The strong rank correlations further show that it reliably recovers how
different latent spaces rank in downstream generation performance.

\paragraph{Separating parallel and orthogonal errors improves prediction.}
To isolate the effect of the directional decomposition, we compare
\sacmetric{} with \sacmetricltwo{}, a basic variant that directly
measures \(\|Y_\alpha-Y_\alpha^\star\|_2\).
As shown in Table~\ref{tab:metric_corr}, even this basic variant correlates more
strongly with downstream gFID than iFID across both model scales.
Decomposing the same error into parallel and orthogonal components further
distinguishes miscalibrated semantic progress from off-direction deviation,
leading to more accurate prediction of downstream generation performance.

\subsection{Improved Tokenizer Diffusability}
\label{sec:tokenizer_diffusability}
We then evaluate whether our improvements to the tokenizer facilitate
downstream diffusion learning. Our experiments compare downstream generation
quality, test whether GSCT and PMSA achieve their intended effects, and assess
the reconstruction and representation quality of the tokenizer itself.

\begin{table*}[t]
\caption{
\textbf{Class-conditional generation on ImageNet \(256\).}
Epochs and Params denote downstream generator training epochs and parameter
count, respectively; a dash denotes an unreported result. Guided results use
the guidance scheme reported by each source. Prior-method values are taken
from RecTok~\citep{shi2026rectok}. Ours achieves
state-of-the-art gFID both without and with guidance.
}
\label{tab:sota_main}
\centering
\setlength{\tabcolsep}{3.5pt}
\resizebox{\textwidth}{!}{
\begin{tabular}{lcccccccccc}
\toprule
\multirow{2}{*}{\textbf{Method}}
& \multirow{2}{*}{\textbf{Epochs}}
& \multirow{2}{*}{\textbf{Params}}
& \multicolumn{4}{c}{\textbf{Without guidance}}
& \multicolumn{4}{c}{\textbf{With guidance}} \\
\cmidrule(lr){4-7}\cmidrule(lr){8-11}
& & &
\textbf{gFID}$\downarrow$ &
\textbf{IS}$\uparrow$ &
\textbf{Prec.}$\uparrow$ &
\textbf{Rec.}$\uparrow$ &
\textbf{gFID}$\downarrow$ &
\textbf{IS}$\uparrow$ &
\textbf{Prec.}$\uparrow$ &
\textbf{Rec.}$\uparrow$ \\
\midrule

\multicolumn{11}{l}{
\textit{Diffusion-centric systems}
} \\
\midrule

DiT~\cite{peebles2023scalable}
& 1400 & 675M
& 9.62 & 121.5 & 0.67 & 0.67
& 2.27 & 278.2 & 0.83 & 0.57 \\

MaskDiT~\cite{zheng2023fast}
& 1600 & 675M
& 5.69 & 177.9 & 0.74 & 0.60
& 2.28 & 276.6 & 0.80 & 0.61 \\

SiT~\cite{ma2024sit}
& 1400 & 675M
& 8.61 & 131.7 & 0.68 & 0.67
& 2.06 & 270.3 & 0.82 & 0.59 \\

MDTv2~\cite{gao2023mdtv2}
& 1080 & 675M
& -- & -- & -- & --
& 1.58 & 314.7 & 0.79 & 0.65 \\

\addlinespace[2pt]

\multirow{2}{*}{REPA~\cite{yu2024representation}}
& 80 & \multirow{2}{*}{675M}
& 7.94 & 121.3 & 0.69 & 0.64
& -- & -- & -- & -- \\
& 800 & {}
& 5.90 & 157.8 & 0.70 & 0.69
& 1.42 & 305.7 & 0.80 & 0.64 \\

\addlinespace[2pt]

\multirow{2}{*}{DDT~\cite{wang2026ddt}}
& 80 & \multirow{2}{*}{675M}
& 6.62 & 135.2 & 0.69 & 0.67
& 1.52 & 263.7 & 0.78 & 0.63 \\
& 400 & {}
& 6.27 & 154.7 & 0.68 & 0.69
& 1.26 & 310.6 & 0.79 & 0.65 \\

\midrule
\multicolumn{11}{l}{
\textit{Tokenizer-centric systems}
} \\
\midrule

\multirow{2}{*}{REPA-E~\cite{leng2025repa}}
& 80 & \multirow{2}{*}{675M}
& 3.46 & 159.8 & 0.77 & 0.63
& 1.67 & 266.3 & 0.80 & 0.63 \\
& 800 & {}
& 1.83 & 217.3 & 0.77 & 0.66
& 1.26 & 314.9 & 0.79 & 0.66 \\

\addlinespace[2pt]

\multirow{2}{*}{VA-VAE~\cite{yao2025reconstruction}}
& 80 & \multirow{2}{*}{675M}
& 4.29 & -- & -- & --
& -- & -- & -- & -- \\
& 800 & {}
& 2.17 & 205.6 & 0.77 & 0.65
& 1.35 & 295.3 & 0.79 & 0.65 \\

\addlinespace[2pt]

AFM~\cite{chen2025aligntok}
& 800 & 675M
& 2.04 & 206.2 & 0.76 & 0.67
& 1.37 & 293.6 & 0.79 & 0.65 \\

\addlinespace[2pt]

SVGTok~\cite{shi2026latent}
& 1400 & 675M
& 3.36 & 181.2 & -- & --
& 1.92 & 264.9 & -- & -- \\

\addlinespace[2pt]

\multirow{2}{*}{RAE~\cite{zheng2026diffusion}}
& 80 & \multirow{2}{*}{839M}
& 2.16 & 214.8 & 0.82 & 0.59
& -- & -- & -- & -- \\
& 800 & {}
& 1.51 & 242.9 & 0.79 & 0.63
& 1.13 & 262.6 & 0.78 & 0.67 \\

\addlinespace[2pt]

\multirow{3}{*}{RecTok~\cite{shi2026rectok}}
& 20 & \multirow{3}{*}{839M}
& 4.44 & 149.0 & -- & --
& -- & -- & -- & -- \\
& 80 & {}
& 2.09 & 198.6 & 0.79 & 0.62
& 1.48 & 223.8 & 0.79 & 0.65 \\
& 600 & {}
& 1.34 & 254.6 & 0.78 & 0.65
& 1.13 & 289.2 & 0.79 & 0.67 \\

\addlinespace[2pt]

\rowcolor{baselinecolor}
& 20 &
& 3.28 & 181.3 & 0.80 & 0.57
& -- & -- & -- & -- \\
\rowcolor{baselinecolor}
\textbf{AffineTok} (ours)
& 80 & 839M
& 1.86 & 217.6 & 0.80 & 0.62
& -- & -- & -- & -- \\
\rowcolor{baselinecolor}
& 600 &
& \textbf{1.21} & \textbf{268.4} & 0.78 & 0.67
& 1.10 & 282.7 & 0.77 & 0.68 \\
\bottomrule
\end{tabular}
}
\end{table*}

\paragraph{Better Diffusability.}
Table~\ref{tab:sota_main} compares our method with RecTok using the same
downstream model and training objective, as well as with other generation
systems.
Our improvements lower gFID throughout training. The benefit is pronounced at
20 epochs, reducing gFID by $26\%$ from $4.44$ to $3.28$ while increasing IS
from $149.0$ to $181.3$.
The improvement persists with extended training, reaching state-of-the-art
gFID of $1.21$ without guidance and $1.10$ with guidance after 600 epochs.

\paragraph{GSCT improves semantic consistency across images.}
We test whether GSCT makes the semantic organization of clean patch latents
more consistent across images. Following~\citet{park2023linear}, we represent
the semantic difference between two classes as the direction between their
latent centroids. If the latent semantics are consistent across images, the
same class-pair direction estimated from different image subsets should agree.
Specifically, we split 20 images from each of 50 ImageNet classes into two
halves, compute class-centroid difference directions in each half, and count
directions whose cross-split angle exceeds a fixed threshold. To reduce
split-induced estimation error, we repeat the analysis over 20 partitions
following~\citet{schutt2023statistical}. As shown in the left panel of
Figure~\ref{fig:semantic-probes}, GSCT reduces the inconsistency rate by $69\%$
on average across all three thresholds, showing that it coordinates the
semantic organization of clean patch latents across images.

\paragraph{PMSA improves semantic recovery from predicted clean latents.}
We test whether PMSA improves the recovery of clean-image semantics from the
clean-latent predictions used in denoising. Following standard probing
practice~\citep{alain2016understanding,hewitt2019designing}, we train a semantic
projector on clean latents, freeze it, and apply it without adaptation to the
predictions of a probe denoiser trained identically for every tokenizer. We
then measure the top-10 semantic retrieval miss rate under severe noise.
As shown in the right panel of Figure~\ref{fig:semantic-probes}, GSCT and PMSA
reduce the miss rate by $56\%$ on average across the three noise levels,
showing that clean-image semantics are recovered more reliably through the
predicted-clean-latent path.

\begin{table*}[t]
    \centering

    \begin{minipage}[t]{0.485\textwidth}
        \vspace{0pt}
        \centering
        \captionsetup{skip=3pt}
    
        \captionof{table}{
            \textbf{Component ablation of GSCT and PMSA.}
            All variants use the same downstream model and training protocol.
        }
        \label{tab:component_ablation}

        \resizebox{0.92\linewidth}{!}{
            \begin{tabular}{lcccc}
                \toprule
                \textbf{Variant}
                & \textbf{gFID}\(\downarrow\)
                & \textbf{IS}\(\uparrow\)
                & \textbf{Prec.}\(\uparrow\)
                & \textbf{Rec.}\(\uparrow\) \\
                \midrule
                RecTok
                & 4.44 & 149.0 & 0.79 & 0.57 \\
                \midrule
                + GSCT
                & 3.67 & 168.38 & 0.80 & 0.56 \\
        
                + PMSA
                & 3.78 & 162.45 & 0.80 & 0.56 \\
        
                \textbf{+ GSCT + PMSA}
                & \textbf{3.28} & \textbf{181.28} & 0.80 & 0.57 \\
                \bottomrule
            \end{tabular}
        }

        \end{minipage}    
    \hfill
    \begin{minipage}[t]{0.485\textwidth}
        \vspace{0pt}
        \centering
        \captionsetup{skip=3pt}

        \captionof{table}{
            \textbf{Role of the global semantic coordination token.}
            REG denotes encoder-side register tokens. Neither CLS nor register
            tokens are provided to the downstream DiT.
        }
        \label{tab:cls_reg_ablation}

        \resizebox{\linewidth}{!}{
            \begin{tabular}{lccccc}
                \toprule
                \textbf{Setting}
                & \textbf{REG target}
                & \textbf{gFID}\(\downarrow\)
                & \textbf{IS}\(\uparrow\)
                & \textbf{Prec.}\(\uparrow\)
                & \textbf{Rec.}\(\uparrow\) \\
                \midrule
                GSCT (CLS)
                & None
                & \textbf{3.67} & \textbf{168.38} & 0.80 & 0.56 \\

                GSCT + 4 REG
                & None
                & 3.69 & 167.75 & 0.80 & 0.56 \\

                GSCT + 4 REG
                & DINOv3 REG
                & 3.84 & 161.84 & 0.80 & 0.57 \\

                4 REG, without GSCT
                & DINOv3 REG
                & 4.06 & 157.55 & 0.80 & 0.57 \\
                \bottomrule
            \end{tabular}
        }
    \end{minipage}
\end{table*}

\paragraph{Improved latent semantics with comparable reconstruction.}
Table~\ref{tab:tokenizer_quality} reports gains of about 10 percentage points
in both pooled and flattened linear-probing accuracy, while reconstruction
quality remains comparable. These results show that our method substantially
improves latent semantics without materially compromising reconstruction.

\subsection{Ablation Studies}
\label{sec:ablation}

We ablate GSCT and PMSA and examine the design of the global semantic coordination
token. All variants use the same downstream model and training protocol and
are evaluated after 20 epochs.

\paragraph{Contributions of GSCT and PMSA.}
Table~\ref{tab:component_ablation} shows that both GSCT and PMSA improve over
RecTok when used alone, while combining them further reduces gFID to $3.28$.
These results are consistent with their complementary roles: GSCT coordinates
the semantic organization of clean patch latents across images, whereas PMSA
directly supervises semantic recovery from predicted clean latents.

\paragraph{Role of global semantic supervision.}
Table~\ref{tab:cls_reg_ablation} examines whether the benefit of GSCT comes
from its global semantic target or merely from adding tokens. Adding four
unsupervised register tokens leaves performance nearly unchanged, while
supervising them with DINOv3 register features degrades it. Using the supervised
register tokens without GSCT performs worse still. These results show that the
global DINOv3 \([\mathrm{CLS}]\) feature provides a shared semantic anchor
across images, rather than merely increasing the number of encoder tokens.

\begin{table*}[t]
\caption{
\textbf{Tokenizer reconstruction and representation quality on ImageNet-1K.}
Following RecTok~\citep{shi2025rectok}, L.P. Acc. denotes top-1 linear-probing
accuracy on frozen latents; \emph{Pooled} averages the patch latents, while
\emph{Flat} vectorizes them. Best results are in bold.
}
\label{tab:tokenizer_quality}
\centering
\scriptsize
\setlength{\tabcolsep}{4pt}
\renewcommand{\arraystretch}{0.95}

\begin{tabular}{lcccccc}
\toprule
\multirow{2}{*}{\textbf{Tokenizer}}
& \multirow{2}{*}{\textbf{rFID}\(\downarrow\)}
& \multirow{2}{*}{\textbf{PSNR}\(\uparrow\)}
& \multirow{2}{*}{\textbf{LPIPS}\(\downarrow\)}
& \multicolumn{2}{c}{\textbf{L.P. Acc.}\(\uparrow\)}
& \multirow{2}{*}{\textbf{\sacmetric{}}\(\downarrow\)} \\
\cmidrule(lr){5-6}
& & & & \textbf{Pooled} & \textbf{Flat} & \\
\midrule
RecTok
& \textbf{0.48}
& \textbf{26.16}
& \textbf{0.1016}
& 62.32
& 58.22
& 52.120 \\

Ours
& 0.49
& 25.83
& 0.1086
& \textbf{72.63}
& \textbf{68.13}
& \textbf{51.495} \\
\bottomrule
\end{tabular}
\end{table*}

%% file: sections/related_work.tex
\section{Related Work}
\label{sec:related_work}

\paragraph{Semantically and structurally organized latent spaces.}
The reconstruction--generation mismatch has motivated visual tokenizers that
shape their latent spaces for downstream generative modeling rather than for
reconstruction alone.
VA-VAE~\citep{yao2025reconstruction} aligns projected spatial latents with
corresponding features from a frozen vision foundation model (VFM) and matches
their within-image pairwise relations.
EQ-VAE~\citep{kouzelis2025eq} encourages latent maps to be equivariant to
anisotropic scaling and rotation through a transformed-latent reconstruction
objective.
SSVAE~\citep{liu2025delving} induces a few-mode bias in the channel
eigenspectrum and a low-frequency bias in the spatio-temporal spectrum.
UAE~\citep{fan2025prism} decomposes clean latent features into a
low-frequency base aligned with frozen semantic-encoder features and
high-frequency residual bands optimized for pixel-faithful reconstruction.
More recently, RecTok~\citep{shi2025rectok} distills VFM semantics into
rectified-flow states obtained by interpolating clean latents with Gaussian
noise.
Send-VAE~\citep{page2026boosting} uses a nonlinear mapper to align similarly
perturbed latents with VFM patch features, aiming to improve semantic
disentanglement.
These methods regularize either clean tokenizer latents or semantic projectors
and mappers applied directly to noisy tokenizer states.
However, they do not constrain the semantics of the posterior-mean
clean latent. We address this component with GSCT, which coordinates clean-latent semantics
across images, and PMSA, which supervises the semantics of posterior-mean
latent estimates.

\paragraph{Metrics for latent diffusability.}
Reconstruction quality does not reliably predict downstream generation
quality.
iFID~\citep{xu2026making} constructs nearest-neighbor latent interpolants and
computes FID between their decoded distribution and real images, measuring
local manifold continuity.
VIV~\citep{zhong2026diffusing} quantifies the irreducible velocity ambiguity
induced by overlapping flow-matching trajectories.
However, neither considers the semantic recovery problem.
By decomposing its error, we identify SAC as a necessary condition and
introduce \sacmetric{}, a tokenizer-side proxy for SAC.

\paragraph{Latent diffusion generation.}
Latent diffusion models (LDMs) reduce the cost of high-resolution synthesis by
learning a denoising process in the compact continuous space of a pretrained
autoencoder~\citep{rombach2022high}.
DiT~\citep{peebles2023scalable} replaces convolutional denoisers with scalable transformers over latent
patches, while SiT~\citep{ma2024sit} places this backbone in a stochastic-interpolant framework
that supports diffusion- and flow-based objectives.
MaskDiT~\citep{zheng2023fast} and MDTv2~\citep{gao2023mdtv2} use masked latent modeling to improve training efficiency and
contextual learning, while DDT~\citep{wang2026ddt} decouples
semantic extraction from velocity decoding.
Beyond architectural design, REPA~\citep{yu2024representation} 
aligns intermediate denoiser states with
clean-image representations from a pretrained visual
encoder.
REG~\citep{wu2026representation} instead entangles low-level
image latents with a high-level class token from a pretrained visual encoder
during denoising, keeping semantic guidance active at inference.
These advances improve the generator after tokenization while leaving the
tokenizer-defined latent geometry fixed. Our work instead studies how that
geometry should organize semantics for downstream denoising.

%% file: sections/conclusion.tex
\section{Conclusion}
\label{sec:conclusion}

In this paper, we revisit a fundamental question: \emph{how should a visual
tokenizer organize semantics to make downstream diffusion easier to learn?}
Existing methods supervise semantics in clean or noisy latents but not those
decoded from diffusion's predicted clean latent. Our exact error decomposition
identifies the missing requirement as \emph{Semantic Affine Consistency} (SAC).
We further operationalize SAC through \sacmetric{}, a tokenizer-side proxy
metric that closely tracks generation quality.
Guided by this analysis, GSCT coordinates semantics across images and
PMSA supervises semantic recovery from posterior-mean latent predictions,
without changing downstream diffusion training. Overall, our tokenizer achieves
gFID \(1.21\) without classifier-free guidance and \(1.10\) with guidance on
ImageNet \(256\times256\). We hope our findings foster further research on
visual tokenizers explicitly organized for downstream generative modeling.

%% file: iclr2026_conference.bib
@article{xu2026making,
  title={Making Reconstruction FID Predictive of Diffusion Generation FID},
  author={Xu, Tongda and He, Mingwei and Abu-Hussein, Shady and Hernandez-Lobato, Jose Miguel and Zheng, Chunhang and Zhao, Kai and Zhou, Chao and Zhang, Ya-Qin and Wang, Yan},
  journal={arXiv preprint arXiv:2603.05630},
  year={2026}
}

@inproceedings{rombach2022high,
  title={High-resolution image synthesis with latent diffusion models},
  author={Rombach, Robin and Blattmann, Andreas and Lorenz, Dominik and Esser, Patrick and Ommer, Bj{\"o}rn},
  booktitle={Proceedings of the IEEE/CVF conference on computer vision and pattern recognition},
  pages={10684--10695},
  year={2022}
}

@article{shi2025rectok,
  title={Rectok: Reconstruction distillation along rectified flow},
  author={Shi, Qingyu and Wu, Size and Bai, Jinbin and Yu, Kaidong and Wang, Yujing and Tong, Yunhai and Li, Xiangtai and Li, Xuelong},
  journal={arXiv preprint arXiv:2512.13421},
  year={2025}
}

@article{alain2016understanding,
  title={Understanding intermediate layers using linear classifier probes},
  author={Alain, Guillaume and Bengio, Yoshua},
  journal={arXiv preprint arXiv:1610.01644},
  year={2016}
}

@inproceedings{hewitt2019designing,
  title={Designing and interpreting probes with control tasks},
  author={Hewitt, John and Liang, Percy},
  booktitle={Proceedings of the 2019 conference on empirical methods in natural language processing and the 9th international joint conference on natural language processing (emnlp-ijcnlp)},
  pages={2733--2743},
  year={2019}
}

@article{schutt2023statistical,
  title={Statistical inference on representational geometries},
  author={Sch{\"u}tt, Heiko H and Kipnis, Alexander D and Diedrichsen, J{\"o}rn and Kriegeskorte, Nikolaus},
  journal={Elife},
  volume={12},
  pages={e82566},
  year={2023},
  publisher={eLife Sciences Publications, Ltd}
}

@article{park2023linear,
  title={The linear representation hypothesis and the geometry of large language models},
  author={Park, Kiho and Choe, Yo Joong and Veitch, Victor},
  journal={arXiv preprint arXiv:2311.03658},
  year={2023}
}

@article{karras2022elucidating,
  title={Elucidating the design space of diffusion-based generative models},
  author={Karras, Tero and Aittala, Miika and Aila, Timo and Laine, Samuli},
  journal={Advances in neural information processing systems},
  volume={35},
  pages={26565--26577},
  year={2022}
}

@article{page2026boosting,
  title={Boosting latent diffusion models via disentangled representation alignment},
  author={Page, John and Niu, Xuesong and Wu, Kai and Gai, Kun},
  journal={arXiv preprint arXiv:2601.05823},
  year={2026}
}

@article{liu2025delving,
  title={Delving into latent spectral biasing of video VAEs for superior diffusability},
  author={Liu, Shizhan and Deng, Xinran and Yang, Zhuoyi and Teng, Jiayan and Gu, Xiaotao and Tang, Jie},
  journal={arXiv preprint arXiv:2512.05394},
  year={2025}
}

@article{kouzelis2025eq,
  title={Eq-vae: Equivariance regularized latent space for improved generative image modeling},
  author={Kouzelis, Theodoros and Kakogeorgiou, Ioannis and Gidaris, Spyros and Komodakis, Nikos},
  journal={arXiv preprint arXiv:2502.09509},
  year={2025}
}

@article{zhong2026diffusing,
  title={Diffusing in the Right Space: A Systematic Study of Latent Diffusability},
  author={Zhong, Tianxiong and Tian, Xingye and Wang, Xuebo and Tao, Xin and Wan, Pengfei},
  journal={arXiv preprint arXiv:2606.03578},
  year={2026}
}

@inproceedings{peebles2023scalable,
  title={Scalable diffusion models with transformers},
  author={Peebles, William and Xie, Saining},
  booktitle={2023 IEEE/CVF International Conference on Computer Vision (ICCV)},
  pages={4172--4182},
  year={2023},
  organization={IEEE}
}

@article{zheng2023fast,
  title={Fast training of diffusion models with masked transformers},
  author={Zheng, Hongkai and Nie, Weili and Vahdat, Arash and Anandkumar, Anima},
  journal={arXiv preprint arXiv:2306.09305},
  year={2023}
}

@inproceedings{ma2024sit,
  title={Sit: Exploring flow and diffusion-based generative models with scalable interpolant transformers},
  author={Ma, Nanye and Goldstein, Mark and Albergo, Michael S and Boffi, Nicholas M and Vanden-Eijnden, Eric and Xie, Saining},
  booktitle={European Conference on Computer Vision},
  pages={23--40},
  year={2024},
  organization={Springer}
}

@article{gao2023mdtv2,
  title={Mdtv2: Masked diffusion transformer is a strong image synthesizer},
  author={Gao, Shanghua and Zhou, Pan and Cheng, Ming-Ming and Yan, Shuicheng},
  journal={arXiv preprint arXiv:2303.14389},
  year={2023}
}

@inproceedings{leng2025repa,
  title={Repa-e: Unlocking vae for end-to-end tuning with latent diffusion transformers},
  author={Leng, Xingjian and Singh, Jaskirat and Hou, Yunzhong and Xing, Zhenchang and Xie, Saining and Zheng, Liang},
  booktitle={2025 IEEE/CVF International Conference on Computer Vision (ICCV)},
  pages={18262--18272},
  year={2025},
  organization={IEEE}
}

@article{yu2024representation,
  title={Representation alignment for generation: Training diffusion transformers is easier than you think},
  author={Yu, Sihyun and Kwak, Sangkyung and Jang, Huiwon and Jeong, Jongheon and Huang, Jonathan and Shin, Jinwoo and Xie, Saining},
  journal={arXiv preprint arXiv:2410.06940},
  year={2024}
}

@inproceedings{yao2025reconstruction,
  title={Reconstruction vs. generation: Taming optimization dilemma in latent diffusion models},
  author={Yao, Jingfeng and Yang, Bin and Wang, Xinggang},
  booktitle={2025 IEEE/CVF Conference on Computer Vision and Pattern Recognition (CVPR)},
  pages={15703--15712},
  year={2025},
  organization={IEEE}
}

@article{chen2025aligntok,
  title={AlignTok: Aligning Visual Foundation Encoders to Tokenizers for Diffusion Models},
  author={Chen, Bowei and Bi, Sai and Tan, Hao and Zhang, He and Zhang, Tianyuan and Li, Zhengqi and Xiong, Yuanjun and Zhang, Jianming and Zhang, Kai},
  journal={arXiv preprint arXiv:2509.25162},
  year={2025}
}

@inproceedings{zheng2026diffusion,
  title={Diffusion transformers with representation autoencoders},
  author={Zheng, Boyang and Ma, Nanye and Tong, Shengbang and Xie, Saining},
  booktitle={International Conference on Learning Representations},
  volume={2026},
  pages={35791--35820},
  year={2026}
}

@inproceedings{shi2026rectok,
  title={Rectok: Reconstruction distillation along rectified flow},
  author={Shi, Qingyu and Wu, Size and Bai, Jinbin and Yu, Kaidong and Wang, Yujing and Tong, Yunhai and Li, Xiangtai and Li, Xuelong},
  booktitle={Proceedings of the IEEE/CVF Conference on Computer Vision and Pattern Recognition},
  pages={40685--40695},
  year={2026}
}

@inproceedings{wang2026ddt,
  title={Ddt: Decoupled diffusion transformer},
  author={Wang, Shuai and Tian, Zhi and Huang, Weilin and Wang, Limin},
  booktitle={Proceedings of the IEEE/CVF conference on computer vision and pattern recognition},
  pages={40633--40642},
  year={2026}
}

@inproceedings{shi2026latent,
  title={Latent diffusion model without variational autoencoder},
  author={Shi, Minglei and Wang, Haolin and Zheng, Wenzhao and Yuan, Ziyang and Wu, Xiaoshi and Wang, Xintao and Wan, Pengfei and Zhou, Jie and Lu, Jiwen},
  booktitle={International Conference on Learning Representations},
  volume={2026},
  pages={154506--154537},
  year={2026}
}

@article{kingma2013auto,
  title={Auto-encoding variational bayes},
  author={Kingma, Diederik P and Welling, Max},
  journal={arXiv preprint arXiv:1312.6114},
  year={2013}
}

@article{van2017neural,
  title={Neural discrete representation learning},
  author={Van Den Oord, Aaron and Vinyals, Oriol and others},
  journal={Advances in neural information processing systems},
  volume={30},
  year={2017}
}

@article{simeoni2025dinov3,
  title={Dinov3},
  author={Sim{\'e}oni, Oriane and Vo, Huy V and Seitzer, Maximilian and Baldassarre, Federico and Oquab, Maxime and Jose, Cijo and Khalidov, Vasil and Szafraniec, Marc and Yi, Seungeun and Ramamonjisoa, Micha{\"e}l and others},
  journal={arXiv preprint arXiv:2508.10104},
  year={2025}
}

@article{russakovsky2015imagenet,
  title={Imagenet large scale visual recognition challenge},
  author={Russakovsky, Olga and Deng, Jia and Su, Hao and Krause, Jonathan and Satheesh, Sanjeev and Ma, Sean and Huang, Zhiheng and Karpathy, Andrej and Khosla, Aditya and Bernstein, Michael and others},
  journal={International journal of computer vision},
  volume={115},
  number={3},
  pages={211--252},
  year={2015},
  publisher={Springer}
}

@article{salimans2016improved,
  title={Improved techniques for training gans},
  author={Salimans, Tim and Goodfellow, Ian and Zaremba, Wojciech and Cheung, Vicki and Radford, Alec and Chen, Xi},
  journal={Advances in neural information processing systems},
  volume={29},
  year={2016}
}

@article{heusel2017gans,
  title={Gans trained by a two time-scale update rule converge to a local nash equilibrium},
  author={Heusel, Martin and Ramsauer, Hubert and Unterthiner, Thomas and Nessler, Bernhard and Hochreiter, Sepp},
  journal={Advances in neural information processing systems},
  volume={30},
  year={2017}
}

@article{kynkaanniemi2019improved,
  title={Improved precision and recall metric for assessing generative models},
  author={Kynk{\"a}{\"a}nniemi, Tuomas and Karras, Tero and Laine, Samuli and Lehtinen, Jaakko and Aila, Timo},
  journal={Advances in neural information processing systems},
  volume={32},
  year={2019}
}

@article{wu2026representation,
  title={Representation entanglement for generation: Training diffusion transformers is much easier than you think},
  author={Wu, Ge and Zhang, Shen and Shi, Ruijing and Gao, Shanghua and Chen, Zhenyuan and Wang, Lei and Chen, Zhaowei and Gao, Hongcheng and Tang, Yao and Cheng, Ming-Ming and others},
  journal={Advances in Neural Information Processing Systems},
  volume={38},
  pages={7714--7743},
  year={2026}
}

@article{fan2025prism,
  title={The prism hypothesis: Harmonizing semantic and pixel representations via unified autoencoding},
  author={Fan, Weichen and Diao, Haiwen and Wang, Quan and Lin, Dahua and Liu, Ziwei},
  journal={arXiv preprint arXiv:2512.19693},
  year={2025}
}
